\documentclass[conference]{IEEEtran}
\IEEEoverridecommandlockouts
\usepackage{cite}
\usepackage{amsmath,amssymb,amsfonts}
\DeclareUnicodeCharacter{2009}{\,} 
\usepackage{algorithmic}
\usepackage{graphicx}
\usepackage{hyperref}
\usepackage{textcomp}
\usepackage{xcolor}
\def\BibTeX{{\rm B\kern-.05em{\sc i\kern-.025em b}\kern-.08em
    T\kern-.1667em\lower.7ex\hbox{E}\kern-.125emX}}

\makeatletter
\def\ps@IEEEtitlepagestyle{%
  \def\@oddfoot{\mycopyrightnotice}%
  \def\@oddhead{\hbox{}\@IEEEheaderstyle\leftmark\hfil\thepage}\relax
  \def\@evenhead{\@IEEEheaderstyle\thepage\hfil\leftmark\hbox{}}\relax
  \def\@evenfoot{}%
}
\def\mycopyrightnotice{%
  \begin{minipage}{\linewidth}
  \scriptsize
  979-8-3503-2391-7/23/\$31.00~\copyright~2023 IEEE
  \end{minipage}
}

\newcommand{\linebreakand}{%
  \end{@IEEEauthorhalign}
  \hfill\mbox{}\par
  \mbox{}\hfill\begin{@IEEEauthorhalign}
}
\let\old@ps@IEEEtitlepagestyle\ps@IEEEtitlepagestyle
\def\confheader#1{%
    \def\ps@IEEEtitlepagestyle{%
        \old@ps@IEEEtitlepagestyle%
        \def\@oddhead{\strut\hfill#1\hfill\strut}%
        \def\@evenhead{\strut\hfill#1\hfill\strut}%
    }%
    \ps@headings%
}
\makeatother

\confheader{%
        \parbox{20cm}{2\textsuperscript{nd} International Conference on Computational Intelligence and Sustainable Engineering Solution (CISES-2023)}
}

\begin{document}

\title{Unified View of Damage  leaves  Planimetry  \&  Analysis  Using Digital Images Processing Techniques\\}
%
%
\author{\IEEEauthorblockN{Pijush Kanti Kumar}
\IEEEauthorblockA{\textit{Department of Information Technology} \\
\textit{Government College of Engineering \& Textile Technology}\\
Serampore, Calcutta, India \\
pijush752000@yahoo.com}
\and
\IEEEauthorblockN{DeepKiran Munjal}
\IEEEauthorblockA{\textit{Department of Computer Applications} \\
\textit{G. L. Bajaj Institute of Technology and Management}\\
Greater Noida, Uttar Pradesh, India\\
deepa.munjal@gmail.com}
\linebreakand
\IEEEauthorblockN{Sunita Rani}
\IEEEauthorblockA{\textit{Department of Engineering Administration} \\
\textit{Noida Institute of Engineering and Technology}\\
Greater Noida, Uttar Pradesh, India\\
Sunita.Tomar2022@gmail.com}
\and
\IEEEauthorblockN{Anurag Dutta}
\IEEEauthorblockA{\textit{Department of Computer Science \& Engineering} \\
\textit{Government College of Engineering \& Textile Technology}\\
Serampore, Calcutta, India \\
anuragdutta.research@gmail.com}
\linebreakand
\IEEEauthorblockN{Liton Chandra Voumik}
\IEEEauthorblockA{\textit{Department of Economics} \\
\textit{Noakhali Science and Technology University}\\
Noakhali, Bangladesh \\
litonvoumik@gmail.com}
\and
\IEEEauthorblockN{A. Ramamoorthy}
\IEEEauthorblockA{\textit{Department of Mathematics} \\
\textit{Velammal Engineering College, Anna University}\\
Chennai, Tamil Nadu, India \\
ramzenithmaths@gmail.com}
}

\maketitle

\begin{abstract}
The detection of  leaf diseases in plants generally involves visual observation of patterns appearing on the leaf surface. However, there are many diseases that are distinguished based on very subtle changes in these visually observable patterns. This paper attempts to identify plant leaf diseases using image processing techniques. The focus of this study is on the detection of citrus leaf canker disease. Canker  is a bacterial infection of leaves. Symptoms of citrus cankers include brown spots on the leaves, often with a watery or oily appearance. The spots (called lesions in botany) are usually yellow. It is surrounded by a halo of the leaves and is found on both the top and bottom of the leaf. This paper describes various methods that have been used to detect citrus leaf canker disease. The methods used are histogram comparison and k-means clustering. Using these methods, citrus canker development was detected based on  histograms generated based on leaf patterns. The results thus obtained can be used, after consultation with experts in the field of agriculture, to identify suitable treatments for the processes used.\\
\end{abstract}

\begin{IEEEkeywords}
\textit{Digital Image Processing, k – Means Clustering, Citrus Leaf Canker Disease}
\end{IEEEkeywords}
\section{Introduction}
In today's world, the farmland volume serves as an additional source of food. Agriculture's productivity has a significant impact on the Indian economy. As a result, it is crucial to identify plant diseases in the field of agriculture. Use of an automatic disease recognition system is advantageous for spotting a plant pathogens in its very early stages. For example in the case, the United States has pine trees that are susceptible to a dangerous disease called little leaf disorder. The affected tree grows slowly and perishes within six years. Parts of the Southern US, including Alabama and Georgia, are affected by it. Early diagnosis in these situations might have been beneficial. The only method currently in use for identifying and detecting \textit{phytopathogens} is professional assessment using only one's unaided eye. This requires a sizable group of specialists and ongoing vegetation surveillance, both of which are very expensive when dealing with large farmlands. Meanwhile, in some nations, farmers lack access to the necessary resources and even the knowledge to speak with experts. Because of this, consulting experts is expensive and time-consuming. The recommended method works well in these circumstances for keeping an eye on huge expanses of harvests. Visual diagnosis of plant diseases is more time-consuming, less accurate, and only practicable in a few locations. However, using an automatic detection method will require less work, less time, and result in a higher degree of accuracy. Brown and yellow spots, early and late scorch, and certain other common bacterial, viral, and fungal diseases are all seen in plants. The size of the diseased area is measured using image processing, and the difference in color of the damaged area is also determined. The method used for breaking down or classifying an image into numerous components is known as image segmentation. Image segmentation can be done in a variety of cases right now, from the straightforward segmentation process to sophisticated color segmentation methods. Usually, these product attributes to elements that people can easily differentiate into separate elements and see. There are numerous techniques for segmenting images because computers lack the ability to recognize objects intelligently.
The image's unique components are the basis for the segmentation algorithm. This could be a section of an image, color features, or boundary details. Diseased leaf spots play an important role in  plant growth environment. Diseases can also be easily identified with the help of affected areas in culture [1]. Usually, of course, leaves clearly show infected areas and are easily identifiable. Therefore, we can say that plant colour change is an essential aspect of  notification. If the crop is in good health, the crop will have different colours, but if the crop dies from some harmful pathogen, it will automatically change colour. Plant diseases affect specific parts. This can lead to decreased productivity. The  main method used in practice to detect plant diseases is early observation by a specialist, exposing the eye. In this feature retrieval-based investigation, influenced leaf parts were examined using the suggested diagonal disagreement of border, color, and appearance variability features. K-Means clustering was employed. Six different disease types were predicted using this method. Results of the performance assessments were eventually achieved. This reviewed work [2]  is based on methods to reduce  computational complexity through improved automated crop disease detection. This is due to the fact that it significantly harms the agricultural industry and makes the disease's symptoms visible. Once more, the possibility for treating and preventing infected plants is clear. In order to detect and categorize plant diseases, it is necessary to look for reliable, affordable, and accurate techniques.  Test photographs of various cotton sheets were initially implemented. The pictures that are are then captured using image processing approaches, and helpful features are extracted for additional analysis. uses a variety of statistical techniques to categorize the images in accordance with the precise issue of the influenced leaf spot regions.  The analysis of the best corresponded to of characteristic conclusions for leaves impacted by the perfect solution was done using the selection of features. In the classification phase [3],  feature values corresponding to the variance of edge, colour and texture features are stored in the image domain. Based on the affected areas of  leaf diseases, the affected sites were identified. 

\section{Materials and Methods}
\subsection{Materials}
The following materials were taken into use. 
\begin{enumerate}
\item Nikon Make 12.5 Megapixels Digital Camera
\item Personal Computer
\item White A4 Paper Sheet for background
\item MATLAB 2013 version or above
\item Photographs of the leaves were arranged in number.
\end{enumerate}
\subsection{Methods}
\subsubsection{Graphical Method}
A leaf with a measurement region was positioned on graph paper with a 1 mm-wide grid. On graph paper, the sheets are meticulously and precisely defined with the aid of a pencil. It was determined how many grids the sheet's detail encompassed in total. The boundary detail is regarded as 1 if it takes up over fifty percent of the grid; otherwise, it is addressed as 0. The precise leaf area is represented by an array of grid numerals.
\subsubsection{Image Processing Method}
A wider audience are capable of determining leaf area thanks to the MATLAB-based technique used for the processing of images, which is a partially automated technique. Code is created using MATLAB 2013 or later. For the latest releases of the application, this code is functional. The procedure is as follows:
\begin{enumerate}
\item Study the picture. 
\item RGB to grayscale conversion. 
\item Create a binary picture from the image that is grayscale. 
\item Do the leaf area calculation.
\end{enumerate}
\section{k Means Clustering}
Data can be grouped in a variety of ways, however the most popular method is the k-Means algorithm [4], which aims to make categories become more comparable while simultaneously maintaining them as far apart as feasible. In essence, distance calculations using Euclidean distance are performed using k-Means. Euclidean distance calculates the distance between two given points $(x_1, y_1)$ and $(x_2, y_2)$ using the following formula 
\begin{equation}
Distance_{euclidian}=\sqrt{(x_2-x_1)^2 + (y_2-y_1)^2}
\end{equation}
The above formula captures  distance in 2D space, but the same holds true in multidimensional space as the number of terms added increases. The '\textit{k}' in \textit{k}-Means represents the number of distinct clusters into which the data is divided. A fundamental caveat [5] of the \textit{k}-Means algorithm is that the data must be continuous in nature. It will not work if your data is categorical in nature.
\subsection{Data Preparation}
As was previously addressed, the majority of methodologies for clustering, including k-Means, rely on the concept of separation. They compute their distance from a specific point and make an effort to reduce it. When distinct variables utilize distinct components, something occurs. For instance, I want to divide the general population of India into different groups since weight is measured in kilograms of weight while heights is measured in centimeters. As can be observed, the variable's components have a significant impact on the matrix of distances above. As a result, standardizing the information prior to seeking clustering activities is an excellent decision.
\subsection{Algorithm}
\textit{k}-Means is an iterative clustering process. This is repeated until an optimal solution or cluster is reached in the problem space. The following pseudo-example covers the basic steps involved in \textit{k}-means clustering, which is commonly used to cluster data.

Start by deciding how many clusters you want, in the present instance three. The \textit{k}-Means algorithm attempts to connect the closest points to the arbitrarily data centers it starts with. 
\begin{figure*}
\centerline{\includegraphics[width = \linewidth]{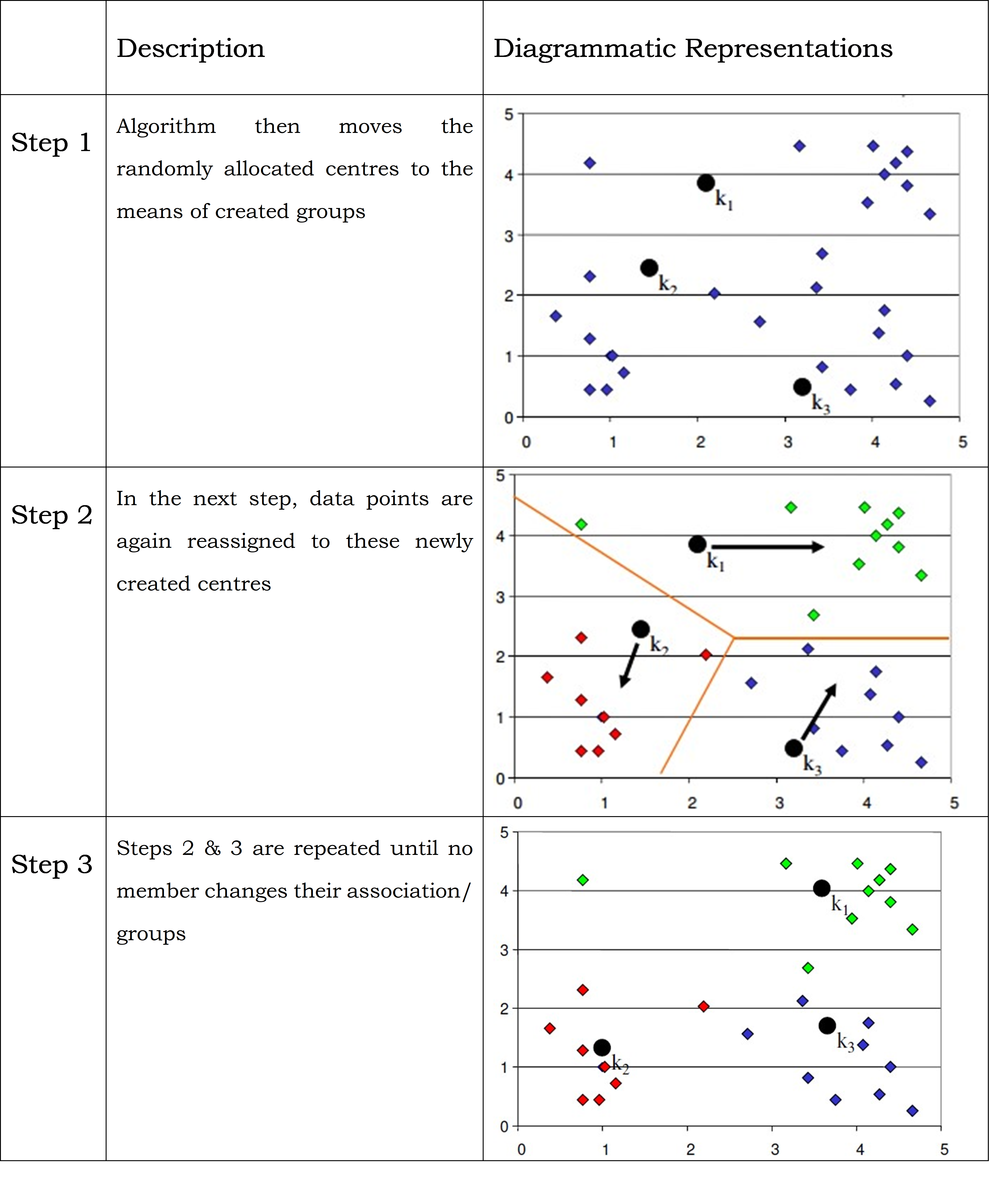}}
\label{fig1}
\caption{Step by Step application of the \textit{k} Means Clustering}
\end{figure*}
\subsection{Image acquisition}
The leaf measuring the affected area  is placed on a black background with no light reflection [6]. Hold the camera  horizontally to the paper. The shooting distance is neither too close nor too far. The photo has been adjusted to cover only the  background. See Fig. 2.
\begin{figure}[htbp]
\centerline{\includegraphics[width = \linewidth]{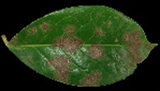}}
\label{fig2}
\caption{Original Image}
\end{figure}
\subsection{Observing the picture}
The image is saved in the computer as $\mathtt{b3.jpeg}$.
This image is read for further processing by MATLAB.
\subsection{Dividing the Image}
The initial picture is divided into two distinct kinds of images in this phase based on how its colors vary. These pictures were chosen for their ability to show impacted and untouched leaf regions. This is done by the \textit{k}-Means clustering method. The algorithm for this method is shown below.
\begin{enumerate}
\item Read the image.
\item Change an image's color space from RGB towards LAB.
\item Use \textit{k}-means clustering to categorize the colors in the 'a*b*' space.
\item With the help of the \textit{k}-means outcomes identify each pixel in the picture.
\item Make pictures that divide the initial picture into sections based on color.
\end{enumerate}
After applying the above algorithm segmented images were obtained. See Fig. 3 and Fig. 4.
\begin{figure}[htbp]
\centerline{\includegraphics[width = \linewidth]{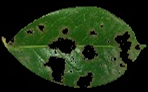}}
\label{fig3}
\caption{Cluster I}
\end{figure}
\begin{figure}[h]
\centerline{\includegraphics[width = \linewidth]{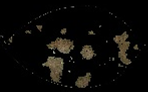}}
\label{fig4}
\caption{Cluster II}
\end{figure}
\subsection{Convert the image (Cluster – I) into Binary image}
After reading the image, I need to first convert this image  from RGB to grayscale [7] and then convert the grayscale image to binary. In this image, the white – coloured areas indicate the unaffected parts of the leaves, as shown in the image in Fig. 5. Useful for calculating the total number of pixels in unaffected areas of the sheet.
\begin{figure}[h]
\centerline{\includegraphics[width = \linewidth]{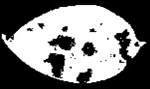}}
\label{fig5}
\caption{Binary Version of Cluster I}
\end{figure}
\subsection{Convert the image (Cluster – II) into Binary image}
After converting the image into binary image, image in Fig. 6 is obtained where white coloured regions indicate affected regions of the leaf. It helps to calculate total number of pixels of affected regions [8] of the leaf. 
\begin{figure}[htbp]
\centerline{\includegraphics[width = \linewidth]{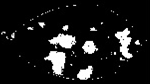}}
\label{fig6}
\caption{Binary Version of Cluster II}
\end{figure}
\subsection{Pixel Calculation}
\subsubsection{From Cluster I}
WP (Pixels of unaffected regions) = 195612 \\
where, WP denotes white pixels of Cluster - I.
\subsubsection{From Cluster II}
WP1 (Pixels of affected regions) = 41246 \\
where, WP1 denotes white pixels of Cluster - II.\\

Now, total pixels of the leaf area that is obtained is shown below. 
\begin{equation}
TP=WP+WP1
\end{equation}
\begin{equation*}
TP=195612+41246=236858
\end{equation*}
Percentage of affected pixels can be obtained by the following as
\begin{equation*}
Error  = \frac{WP1\ (pixels\ of\ affected\ regions)}{TP\ (Total\ pixels\ of\ the\ leaf\ area)} \times 100
\end{equation*}
ERROR (\%) = 17.4138 \%
\section{Results and Discussion}
Leaves were selected from different plots and different types of citrus leaves to test the performance [9] of the new measurement system. Leaf area is calculated using the square grid method and is considered  standard area. When we obtained sampled data from different affected leaves. Then you can get a clear idea about the whole damaged sheet in relation to the specific area. This proposed method is faster [10] and more accurate  than any standard method.
\section{Conclusion}
This article discusses a method for measuring the area of citrus leaves using the processing of digital images. Calculating the quantity of leaves that are damaged has been demonstrated to be possible with the executed algorithmic structure. Although more time-consuming, grid calculating strategies are highly precise for measuring the surface area of leaves. The image processing method is also a fast method [11 - 21] with high  accuracy [22] and precision [23]. You may modify the leaf picture at any moment if you solely preserve it [24]. This can serve as an empirical basis for creating an affordable leaf area meter that meets precision farming requirements. For precise area computations and disease incidence projections for chemical pesticides and application of fertilizer, the proportion of error aspect is crucial and required.

\end{document}